\documentclass[runningheads]{llncs}
\usepackage{graphicx}
\usepackage{multirow, multicol}
\usepackage{booktabs}
\usepackage{hyperref}
\usepackage{xcolor}
\usepackage{makecell}
\usepackage{amsmath,amssymb,amsfonts}
\begin{document}
\title{UM-CAM: Uncertainty-weighted Multi-resolution Class Activation Maps for Weakly-supervised Fetal Brain Segmentation}

%
\titlerunning{Weakly-supervised Fetal Brain Segmentation}
%
\author{Jia Fu\inst{1} \and
Tao Lu\inst{2} \and
Shaoting Zhang\inst{1,3} \and 
Guotai Wang\inst{1,3}}
\authorrunning{J. Fu et al.}
%
\institute{University of Electronic Science and Technology of China, Chengdu, 611731, China \and
Sichuan Provincial People’s Hospital, Chengdu, 610072, China \and
Shanghai Artificial Intelligence Laboratory, Shanghai, 200030, China}
\maketitle              
\begin{abstract}
Accurate segmentation of the fetal brain from Magnetic Resonance Image (MRI) is important for prenatal assessment of fetal development. Although deep learning has shown the potential to achieve this task, it requires a large fine annotated dataset that is difficult to collect. To address this issue, weakly-supervised segmentation methods with image-level labels have gained attention, which are commonly based on class activation maps from a classification network trained with image tags. However, most of these methods suffer from incomplete activation regions, due to the low-resolution localization without detailed boundary cues. To this end, we propose a novel weakly-supervised method with image-level labels based on semantic features and context information exploration. We first propose an Uncertainty-weighted Multi-resolution Class Activation Map (UM-CAM) to generate high-quality pixel-level supervision. Then, we design a Geodesic distance-based Seed Expansion (GSE) method to provide context information for rectifying the ambiguous boundaries of UM-CAM. Extensive experiments on a fetal brain dataset show that our UM-CAM can provide more accurate activation regions with fewer false positive regions than existing CAM variants, and our proposed method outperforms state-of-the-art weakly-supervised methods with image-level labels.

\end{abstract}

\section{Introduction}
Brain extraction is the first step in fetal Magnetic Resonance Image (MRI) analysis in advanced applications such as brain tissue segmentation~\cite{makropoulos2014automatic} and quantitative measurement~\cite{sridar2016automatic,shi2020fetal}, which is essential for assessing fetal brain development and investigate the neuroanatomical correlation of cognitive impairments~\cite{makropoulos2018review}. Current research based on Convolutional Neural Network (CNN)~\cite{ebner2020automated,salehi2018real} has achieved promising performance for automatic fetal brain extraction from pixel-wise annotated fetal MRI. However, it is labor-intensive, time-consuming, and expensive to collect a large-scale pixel-wise annotated dataset, especially for images with poor quality and large variations. To address these issues, weakly-supervised segmentation methods with image-level supervision~\cite{shen2023survey} are introduced due to their minimal annotation demand. However, learning from image-level supervision is extremely challenging since the image-level label only provides the existence of object class, but cannot indicate the information about location and shape that are essential for the segmentation task~\cite{ahn2018learning}.

Prevailing methods learning from image-level labels for segmentation commonly produce a coarse localization of the objects based on Class Activation Maps (CAM)~\cite{zhou2016learning}. Due to the weak annotation, the CAMs from the classification network can only provide rough localization and coarse boundaries of objects. To alleviate the problem, a lot of approaches have been proposed, which can be categorized as one-stage and two-stage methods. One-stage methods aim to generate pixel-level segmentation by training a segmentation branch simultaneously with a classification network. For example, Reliable Region Mining (RRM)~\cite{zhang2020reliability} comprises two parallel branches, in which pixel-level pseudo masks are produced from the classification branch and refined by Conditional Random Field (CRF) to supervise the segmentation branch. Despite their efficiency, one-stage methods commonly achieve inferior segmentation accuracy and incomplete activation of targets, owing to the failure to capture detailed contextual information from image-level labels~\cite{araslanov2020single,gao2021ts}.

In contrast to one-stage methods, two-stage methods can perform favourably, as they leverage dense labels generated by the classification network to train a segmentation network~\cite{li2021pseudo}. For instance, Discriminative Region Suppression (DRS)~\cite{kim2021discriminative} suppresses the attention on discriminative regions and expands it to adjacent less activated regions. However, these methods leverage the CAMs from the deep layer of the classification network and raise the inherent drawback, i.e., low resolution, leading to limited localization and smooth boundaries of objects. Han et al.~\cite{han2022multi} proposed multi-layer pseudo supervision to reduce the false positive rate in segmentation results, while the weights for pseudo masks from different layers are constants that cannot be adaptive. Besides, though the quality of CAMs improves, they are still insufficient to provide accurate object boundaries for segmentation. Numerous methods~\cite{fan2020learning,kolesnikov2016seed,lee2021railroad} have been proposed to explore boundary information. For example, Kolesnikov et al.~\cite{kolesnikov2016seed} proposed a joint loss function that constrains the global weighted rank pooling and low-level object boundary to expand activation regions. AffinityNet~\cite{ahn2018learning} trains another network to learn the semantic similarity between pixels and then propagates the semantics to adjacent pixels via random walk. Nevertheless, these methods use the initial seeds generated from the CAM method, resulting in limited performance when the object-related seeds from CAM are small and sparse. Thus, improving the initial prediction and exploring boundary information are both important for accurate object segmentation. 

In this work, we propose a novel weakly-supervised method for accurate fetal brain segmentation using image-level labels. Our contribution can be summarized as follows: 1) We design an Uncertainty-weighted Multi-resolution CAM (UM-CAM) to integrate low- and high-resolution CAMs via entropy weighting, which can leverage the semantic features extracted from the classification network adaptively and eliminate the noise effectively. 2) We propose a Geodesic distance-based Seed Expansion (GSE) method to generate Seed-derived Pseudo Labels (SPLs) that can provide boundary cues for training a better segmentation model. 3) Extensive experiments conducted on a fetal brain dataset demonstrate the effectiveness of the proposed method, which outperforms several state-of-the-art approaches for learning from image-level labels. 

\begin{figure}[t]
\includegraphics[width=\textwidth]{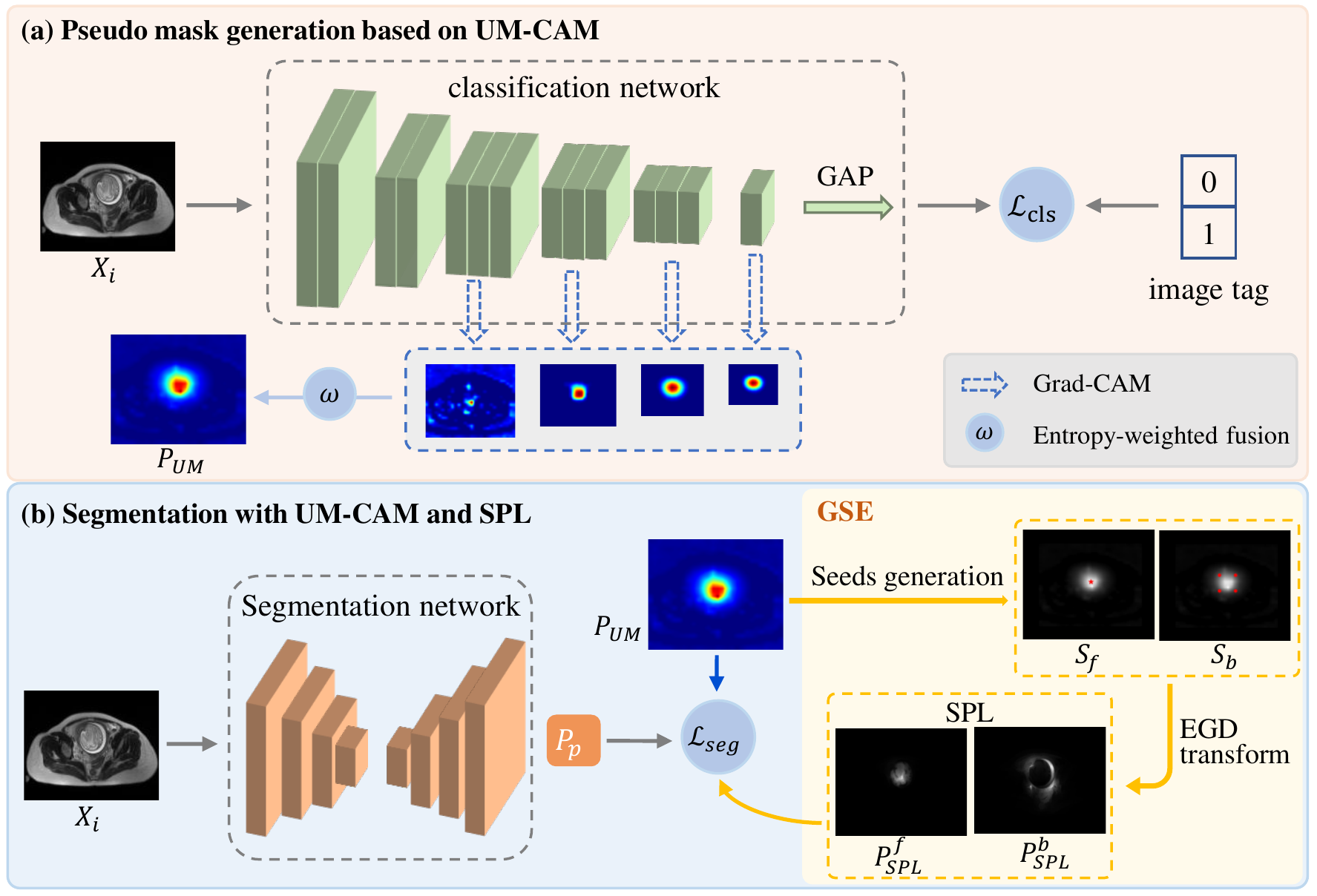}
\caption{Overview of the proposed method. (a) Uncertainty-weighted Multi-resolution CAM (UM-CAM) obtained by a classification network, (b) Segmentation model trained with UM-CAM and auxiliary supervision from Seed-derived Pseudo Label (SPL).}
\label{fig: framework}
\end{figure}

\section{Method}
An overview of our method is presented in Fig.~\ref{fig: framework}. First, to obtain high-quality pseudo masks, Uncertainty-weighted Multi-resolution CAM (UM-CAM) is produced by fusing low- and high-resolution CAMs from different layers of the classification network. Second, seed points are obtained from the UM-CAM automatically, and used to generate Seed-derived Pseudo Labels (SPL) via Geodesic distance-based Seed Expansion (GSE). The SPL provides more detailed context information in addition to UM-CAM for training the final segmentation model.

\subsection{Psuedo Mask Generation Based on UM-CAM}
\subsubsection{Initial Response via Grad-CAM.}
A typical classification network consists of convolutional layers as a feature extractor, followed by global average pooling and a fully connected layer as the output classifier~\cite{chang2020weakly}. Given a set of training images, the classification network is trained with class labels. After training, the Grad-CAM method is utilized to compute the weights $\alpha_{k}$ for the $k$-th channel of a feature map $f$ at a certain layer via gradient backpropagation from the output node for the foreground class. The foreground activation map $A$ can be obtained from a weighted combination of feature maps and followed by a ReLU activation~\cite{selvaraju2017grad}, which is formulated as:
\begin{equation}
\alpha_{k} = \frac{1}{N}\sum_{i=1}^{N} \frac{\partial y}{\partial f_{k}(i)}, 
\label{eq_weight_score}
\end{equation}

\begin{equation}
A(i) = ReLU(\sum_{k}^{} \alpha_{k} f_{k}(i)), 
\label{eq_GradCAM}
\end{equation}
where $y$ is classification prediction score for the foreground. $i$ is pixel index, and $N$ is the pixel number in the image.

\subsubsection{Multi-resolution Exploration and Integration.}
The localization map for each image typically provides discriminative object parts, which is insufficient to provide supervision for the segmentation task. As shown in Fig.~\ref{fig: framework}(a), the activation maps generated from the shallow layers of the classification network contain high-resolution semantic features but suffer from noisy and dispersive localization. In contrast, activation maps generated from the deeper layers perform smoother localization but lack high-resolution information. To take advantage of activation maps from shallow and deep layers, we proposed UM-CAM to integrate the multi-resolution CAMs by uncertainty weighting. Let us denote a set of activation maps from $M$ convolutional blocks as $\mathcal{A}=\{A_{m}\}_{m=0}^{M}$. Each  activation map 
is interpolated to the input size and normalized by its maximum to the range of [0,1], and the normalized activation maps are $\hat{\mathcal{A}}=\{\hat{A}_{m}\}_{m=0}^{M}$. To minimize the uncertainty of pseudo mask, UM-CAM integrates the confident region of multi-resolution CAMs adaptively, which can be presented as the entropy-weighted combination of CAMs:
\begin{equation}
w_{m}(i) = 1 - (-\sum_{j=(b,f)} \hat{A}_m^j(i)log \hat{A}_m^j(i)),
\label{eq: weight_layercam}
\end{equation}

\begin{equation}
P_{UM}(i) = \frac{\sum_{m}w_{m}(i)\hat{A}_{m}(i)}{\sum_{m}w_{m}(i)},
\label{eq_layers}
\end{equation}
where $\hat{A}_m^b$ and $\hat{A}_m^f$ represent the background and foreground probability of $A_m$, respectively. $w_{m}$ is the weight map for $\hat{A}_{m}$, and $P_{UM}$ is the UM-CAM for the target.

\subsection{Robust Segmentation with UM-CAM and SPL}
Though UM-CAM is better than the CAM from the deep layer of the classification network, it is still insufficient to provide accurate object boundaries that are important for segmentation. Motivated by~\cite{luo2021mideepseg}, we propose a Geodesic distance-based Seed Expansion (GSE) method to generate Seed-derived Pseudo Label (SPL) that contains more detailed context information. The SPL is combined with UM-CAM to supervise the segmentation model, as shown in Fig.~\ref{fig: framework} (b).

Concretely, we adopt the centroid and the corner points of the bounding box obtained from UM-CAM as the foreground seeds $S_f$ and background seeds $S_b$, respectively. To efficiently leverage these seed points, SPL is generated via Exponential Geodesic Distance (EGD) transform of the seeds, leading to a foreground cue map $P_{SPL}^f$ and a background cue map $P_{SPL}^b$. The values of $P_{SPL}^f$ and $P_{SPL}^b$ represent the similarity between each pixel and background/foreground seed points, which can be computed as:

\begin{equation}
P_{SPL}^b (i) = e^{-\alpha \cdot D_{b}(i)}, \quad P_{SPL}^f (i) = e^{-\alpha \cdot D_{f}(i)}, 
\label{eq: EGDM}
\end{equation}

\begin{equation}
D_{b}(i) = \min_{j\in S_b} D_{geo}(i,j,I), \quad D_{f}(i) = \min_{j\in S_f} D_{geo}(i,j,I),
\end{equation}

\begin{equation}
D_{geo}\left ( i,j,I \right ) =\min_{p\in P_{i,j}}\int_{0}^{1}\left \| \bigtriangledown I\left ( p\left ( n \right )  \right ) \cdot u\left ( n \right )  \right \|dn
\label{eq_GDM}
\end{equation}
where $P_{i,j}$ is the set of all paths between pixels $i$ and $j$. $D_{b}(i)$ and $D_{f}(i)$ represent the minimal geodesic distance between target pixel $i$ and background/foreground seed points, respectively. $p$ is one feasible path and it is parameterized by $n\in \left [ 0,1 \right ]$. $u\left ( n \right ) = {p}'\left ( n \right ) / \left \| {p}' \left ( n \right ) \right \|$ is a unit vector that is tangent to the direction of the path.

Based on the supervision from UM-CAM and SPL, the segmentation network can be trained by minimizing the following joint object function:
\begin{equation}
L_{seg} = \lambda L_{CE}(P_p, P_{UM}) + (1-\lambda) L_{CE}(P_p, P_{SPL}).
\label{eq: total_loss}
\end{equation}
where $P_p$ is the prediction of the segmentation network, and $\lambda$ is a weight factor to balance the supervision of UM-CAM and SPL. $L_{CE}$ is the Cross-Entropy (CE) loss.

\section{Experiments and results}
\subsection{Experimental Details}
\subsubsection{Dataset.} We collected clinical T2-weighted MRI data of 115 pregnant women in the second trimester with Single-shot Fast Spin-echo (SSFSE). The data were acquired in axial view with pixel size between 0.5547 mm $\times$ 0.5547 mm and 0.6719 mm $\times$ 0.6719 mm and slice thickness between 6.50 mm and 7.15 mm. Each slice was resampled to a uniform pixel size of 1 mm $\times$ 1 mm. In all experiments, we used 80 volumes with 976 positive and 3140 negative slices for training, 10 volumes with 116 positive and 408 negative slices for validation, and 25 volumes with 318 positive and 950 negative slices for testing. Positive and negative slice mean containing the brain and not, respectively. The ground truth was manually annotated by radiologists. Note that we used image-level labels for training and pixel-level ground truth for validation and testing.

\subsubsection{Implementation Details.} To boost the generalizability, we applied spatial and intensity-based data augmentation during the training stage, including gamma correction, random rotation, random flipping, and random cropping. For 2D classification, we employed VGG-16~\cite{simonyan2014very} as backbone architecture, in which an additional convolutional layer is used to substitute the last three fully connected layers. The classification network was trained with 200 epochs using CE loss. Stochastic Gradient Descent (SGD) algorithm was used for optimization with batch size 32, momentum 0.99, and weight decay $5 \times 10^{-4}$. The learning rate was initialized to $1 \times 10^{-3}$. For 2D segmentation, UNet~\cite{ronneberger2015u} was adopted as the backbone architecture. The learning rate was set as 0.01, and the SGD optimizer was trained for 200 epochs with batch size 12, momentum 0.9, and weight decay $5 \times 10^{-4}$. The hyper-parameter $M$ and $\lambda$ were set as 4 and 0.1 based on the best results on the validation set, respectively. We used Dice Similarity Coefficient (DSC) and 95\% Hausdorff Distance ($HD_{95}$) to evaluate the quality of 2D pseudo masks and the final segmentation results in 3D space.

\begin{figure*}[t]
\centerline{\includegraphics[width=\textwidth]{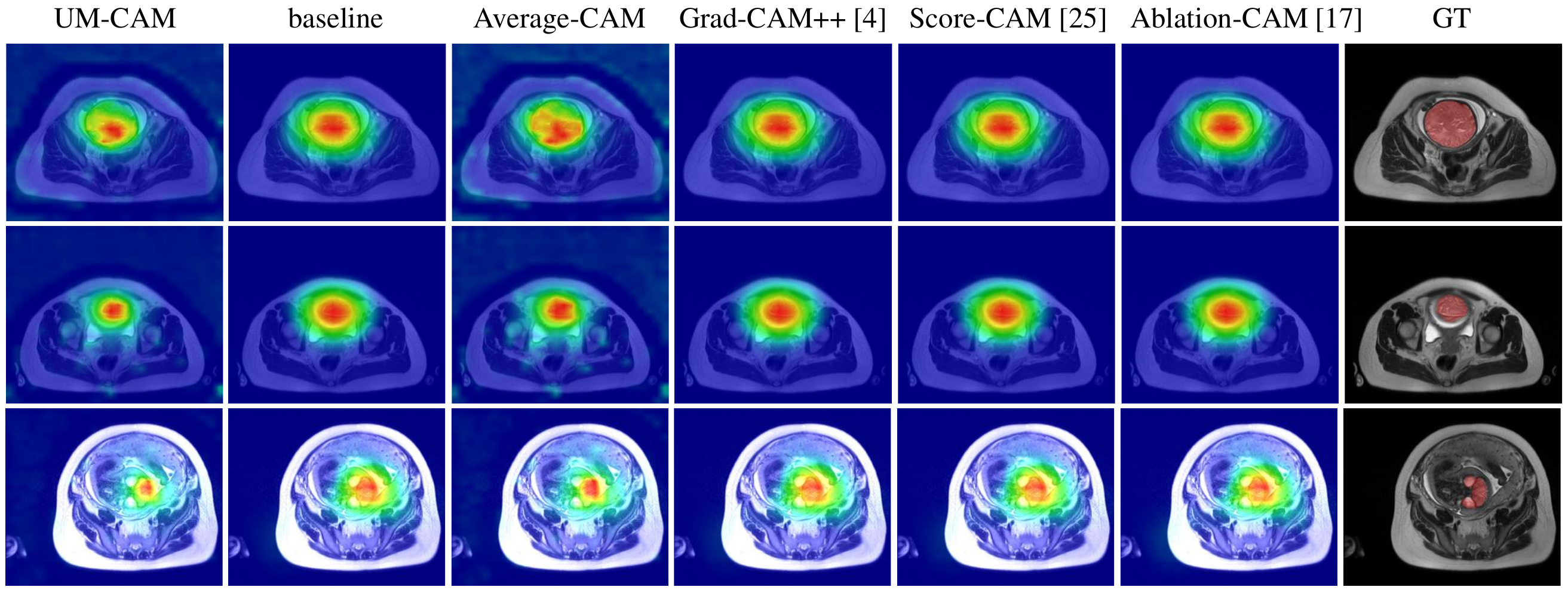}}
\caption{Visual comparison between CAMs obtained by different methods.}
\label{fig: vis_pseudo}
\end{figure*}

\begin{table}[t]
  \centering
  \caption{Ablation study on the validation set to validate the effectiveness of UM-CAM and SPL. * denotes $p$-value $<$ 0.05 when comparing with the second place method.}
    \begin{tabular}{cccc}
    \toprule
    \multicolumn{2}{c}{Method} & DSC (\%) & ~$HD_{95}$ (pixels)~ \\
    \midrule
    \multicolumn{1}{c}{\multirow{3}[2]{*}{\makecell[c]{Pseudo mask \\generation}}} & ~Grad-CAM (baseline)~ & 74.48±13.26 & 39.88±14.66 \\
          & Average-CAM & 77.40±13.28 & 38.70±19.75 \\
          & UM-CAM & \textbf{79.00±12.83*~} & \textbf{35.33±16.98} \\
    \midrule
    \multicolumn{1}{c}{\multirow{4}[2]{*}{Segmentation}}
          & Grad-CAM (baseline)   & 78.69±10.02 & 22.17±16.60 \\
          & UM-CAM & 85.22±6.62 & 5.26±4.23 \\
          & SPL   & 89.05±4.30 & 3.84±4.07 \\
          & UM-CAM+SPL & \textbf{89.76±5.09*~} & \textbf{3.10±2.61} \\
    \bottomrule
    \end{tabular}%
  \label{tab: ablation}%
\end{table}%

\subsection{Ablation Studies}
\subsubsection{Stage1: Quality of Pseudo Masks Obtained by UM-CAM.}
To evaluate the effectiveness of UM-CAM, we compared different pseudo mask generation strategies: 1) Grad-CAM (baseline): only using CAMs from the last layer of the classification network generated by using Grad-CAM method~\cite{selvaraju2017grad}, 2) Average-CAM: fusing multi-resolution CAMs via averaging, 3) UM-CAM: fusing multi-resolution CAMs via uncertainty weighting. Table~\ref{tab: ablation} lists the quantitative evaluation results of these methods, in which the segmentation is converted from CAMs using the optimal threshold found by grid search method. It can be seen that when fusing the information from multiple convolutional layers, the quality of pseudo masks improves. The proposed UM-CAM improves the average DSC by 4.52\% and 1.60\% compared with the baseline and Average-CAM. Fig.~\ref{fig: vis_pseudo} shows a visual comparison between CAMs obtained by the different methods. It can be observed that there are fewer false positive activation regions of UM-CAM compared with the other methods.

\begin{figure*}[t]
\centerline{\includegraphics[width=\textwidth]{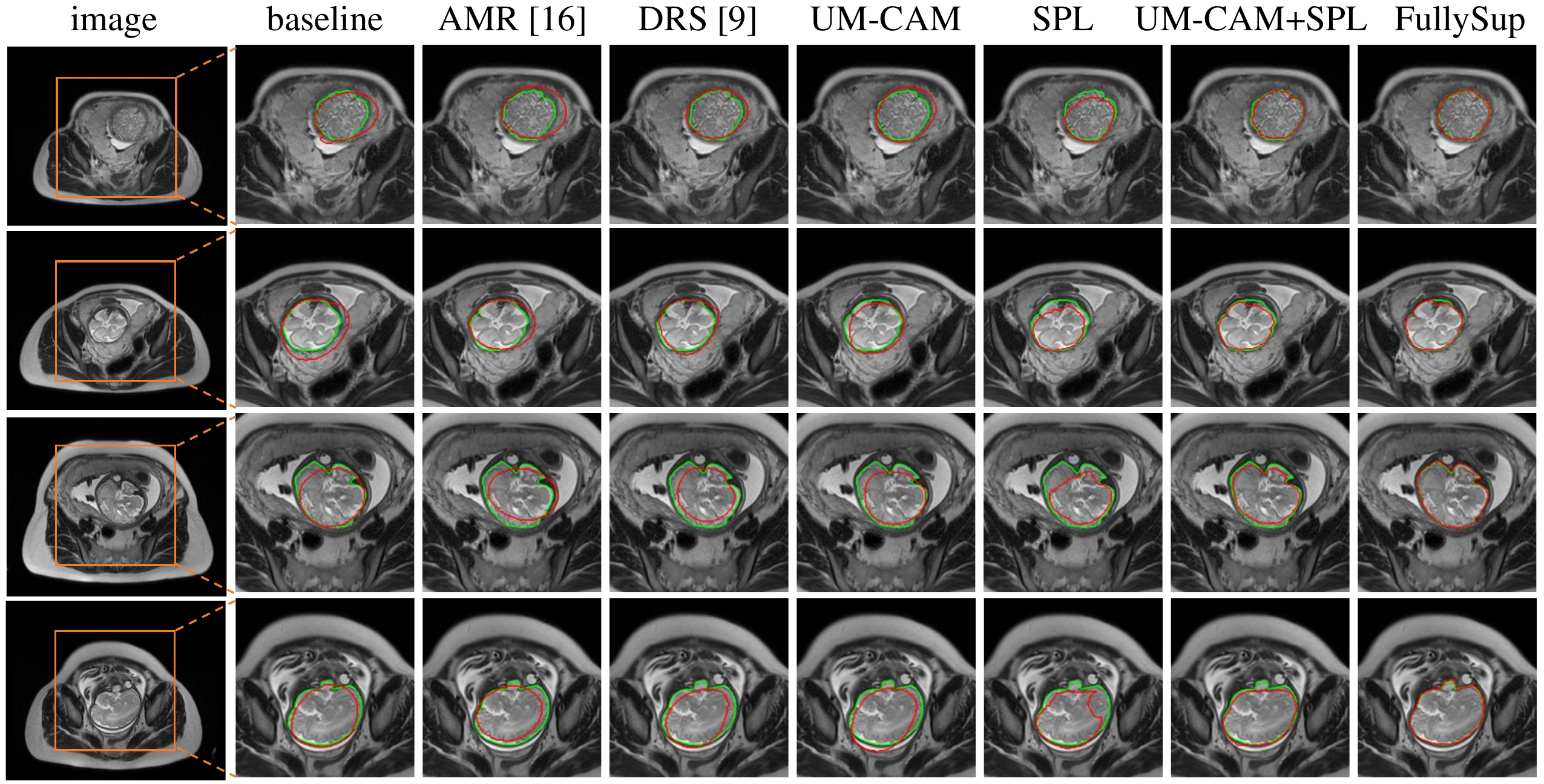}}
\caption{Visual comparison of our method and other weakly-supervised segmentation methods. The green and red contours indicate the boundaries of ground truths and segmentation results, respectively.}
\label{fig: vis_seg}
\end{figure*}

\subsubsection{Stage2: Training Segmentation Model with UM-CAM and SPL.}
To investigate the effectiveness of SPL, we compared it with several segmentation models: 1) Grad-CAM (baseline): only using the pseudo mask generated from Grad-CAM to train the segmentation model, 2) UM-CAM: only using UM-CAM as supervision for the segmentation model, 3) SPL: only using SPL as supervision, 4) UM-CAM+SPL: our proposed method using UM-CAM and SPL supervision for the segmentation model. Quantitative evaluation results in the second section of Table~\ref{tab: ablation} show that the network trained with UM-CAM and SPL supervision achieves an average DSC score of 89.76\%, improving the DSC by 11.07\% compared to the baseline model. Fig.~\ref{fig: vis_seg} depicts a visual comparison between these models. It shows that SPL supervision with context information can better discriminate the fetal brain from the background, leading to more accurate boundaries for segmentation.

\subsection{Comparison with State-of-the-art Methods}
We compared CAM invariants with our UM-CAM in pseudo mask generation stage, such as Grad-CAM$++$~\cite{chattopadhay2018grad}, Score-CAM~\cite{wang2020score}, and Ablation-CAM~\cite{ramaswamy2020ablation}. Table~\ref{tab: sota} shows the quantitative results of these CAM variants. It can be seen that GradCAM$++$, Score-CAM, and Ablation-CAM achieve similar performance, which is consistent with the visualization results shown in Fig.~\ref{fig: vis_pseudo}. The proposed UM-CAM achieves higher accuracy than existing CAM variants, which generates more accurate boundaries that are closer to the ground truth.

\begin{table}[t]
  \centering
  \caption{Comparison between ours and existing weakly-supervised segmentation methods. * denotes $p$-value $<$ 0.05 when comparing with the second place weakly supervised method.}
    \begin{tabular}{ccccc}
    \toprule
    \multirow{2}[4]{*}{Method} & \multicolumn{2}{c}{Validation set} & \multicolumn{2}{c}{Test set} \\
\cmidrule{2-5}          & DSC (\%) & $HD_{95}$ (pixels) & DSC (\%) & $HD_{95}$ (pixels) \\
    \midrule
    Grad-CAM++~\cite{chattopadhay2018grad} & 74.52±13.29 & 39.87±14.69 & 76.60±10.58 & 37.49±11.72 \\
    Score-CAM~\cite{wang2020score} & 74.49±13.35 & 39.88±14.83 & 76.58±10.60 & 37.54±11.73 \\
    Ablation-CAM~\cite{ramaswamy2020ablation} & 74.56±13.29 & 39.76±14.64 & 76.55±10.61 & 37.57±11.82 \\
    AMR~\cite{qin2022activation}   & 78.77±8.83 & 11.53±9.82 & 79.79±6.86 & 11.35±8.02 \\
    DRS~\cite{kim2021discriminative}   & 84.98±5.62 & 7.17±8.01 & 83.79±7.81 & 7.06±5.13 \\
    UM-CAM+SPL (ours) & \textbf{~89.76±5.09*} & \textbf{3.10±2.61*} & \textbf{90.22±3.75*} & \textbf{4.04±4.26*} \\
    \midrule
    FullySup & 95.98±3.17 & 1.22±0.65 & 96.51±2.67 & 1.10±0.40 \\
    \bottomrule
    \end{tabular}%
  \label{tab: sota}%
\end{table}%

We further compared the proposed method with fully supervised method (FullySup) and two weakly-supervised methods, including DRS~\cite{kim2021discriminative} that spreads the attention to adjacent non-discriminative regions by suppressing the attention on discriminative regions and AMR~\cite{qin2022activation} that incorporates a spotlight branch and a compensation branch to dig out more complete object regions. Table~\ref{tab: sota} lists the segmentation results of these methods. Our proposed method achieves an average DSC of 90.22\% and an average $HD_{95}$ of 4.04 pixels, which is at least 3.02 pixels lower than other weakly-supervised methods on the testing set. It indicates that the proposed method can generate segments with more accurate boundaries. Fig.~\ref{fig: vis_seg} shows some qualitative visualization results. The DRS and AMR predictions appear to be coarse and inaccurate in the boundary regions, while our proposed method generates more accurate segmentation results, even similar to those generated from the fully supervised model for some easy samples.

\section{Conclusion}
In this paper, we presented an uncertainty and context-based method for fetal brain segmentation using image-level supervision. An Uncertainty-weighted Multi-resolution CAM (UM-CAM) was proposed to integrate multi-resolution activation maps via uncertainty weighting to generate high-quality pixel-wise supervision. We proposed a Geodesic distance-based Seed Expansion (GSE) method to produce Seed-derived Pseudo Labels (SPL) containing detailed context information. The SPL is combined with UM-CAM for training the segmentation network. The proposed method was evaluated on the fetal brain segmentation task, and experimental results demonstrated the effectiveness of the proposed method and suggested the potential of our proposed method for obtaining accurate fetal brain segmentation with low annotation cost. In the future, it is of interest to validate our method with other segmentation tasks and apply it to other backbone networks.

\subsubsection{Acknowledgement.}
This work was supported by the National Natural Science Foundation of China (62271115).

\bibliographystyle{splncs04}
\bibliography{ref.bib}

\end{document}